\documentclass[letter, 10 pt, conference]{ieeeconf}  

\IEEEoverridecommandlockouts                              

\usepackage[T1]{fontenc}
\usepackage{graphics} 
\usepackage{epsfig} 
\usepackage[linesnumbered,ruled]{algorithm2e}
\usepackage{subcaption}
\usepackage{float}
\usepackage{graphicx}
\usepackage{bm}
\usepackage{balance}
\usepackage{amsmath}
\usepackage{newtxtext}
\usepackage{newtxmath}  
\usepackage{breqn}
\usepackage{cite}
\usepackage{threeparttable}
\usepackage{multirow}
\usepackage{booktabs}
\usepackage{bigstrut}
\usepackage{makecell}
\usepackage{color,soul}
\definecolor{gray}{rgb}{0.5,0.5,0.5}
\usepackage{xspace}
\usepackage{ifthen}
\usepackage[normalem]{ulem} 

\let\origtextcolor\textcolor
\renewcommand{\textcolor}[2]{%
  \ifthenelse{\equal{#1}{red}\OR\equal{#1}{blue}}%
    {\origtextcolor{black}{#2}}%
    {\origtextcolor{#1}{#2}}%
}

\newif\iftracked
\trackedfalse 
\DeclareRobustCommand{\add}[1]{\iftracked\origtextcolor{blue}{#1}\else#1\fi}
\DeclareRobustCommand{\del}[1]{}
\newcommand{\addblk}[1]{\iftracked\origtextcolor{blue}{#1}\else#1\fi}

\newcommand{\model}{{RCM-ACT}\xspace}

\title{
RCM-ACT: Imitation Learning with Dynamic RCM Calibration for Autonomous Intraocular Foreign Body Removal
}

\author{Yue Wang$^{1\dag}$, Wenjie Deng$^{1\dag}$, Haotian Xue$^{2}$, Di Cui$^2$, Yiqi Chen$^{3}$,
\\Mingchuan Zhou$^{4*}$, Haochao Ying$^{5*}$, and Jian Wu$^6$
\thanks{
$^{1}$Yue Wang and Wenjie Deng are with the College of Computer Science and Technology, Zhejiang University, Hangzhou 310012, China. They are also with the State Key Laboratory of Transvascular Implantation Devices of the Second Affiliated Hospital, Zhejiang University School of Medicine, Hangzhou 310009, China, and Zhejiang Key Laboratory of Medical Imaging Artificial Intelligence, Hangzhou 310058, China. (E-mail: ywang2022@zju.edu.cn)}
\thanks{$^{2}$Haotian Xue and Di Cui are with the Dessight Biomedical, Hangzhou 310023, China.}
\thanks{$^{3}$Yiqi Chen is with the Center for Rehabilitation Medicine, Department of Ophthalmology, Zhejiang Provincial People’s Hospital (Zhejiang Key Laboratory of Precision Medicine for Eye Diseases \& Affiliated People’s Hospital, Hangzhou Medical College), Hangzhou 310014, China.}
\thanks{$^{4}$Mingchuan Zhou is with the School of Biosystems Engineering and Food Science, Zhejiang University, Hangzhou 310009, China.}
\thanks{$^{5}$Haochao Ying is with the School of Public Health and Second Affiliated Hospital, Zhejiang University School of Medicine, Hangzhou 310058, China. (E-mail: haochaoying@zju.edu.cn)}
\thanks{$^{6}$Jian Wu is with the State Key Laboratory of Transvascular Implantation Devices of the Second Affiliated Hospital and School of Public Health, Zhejiang University School of Medicine, Hangzhou 310009, China. He is also with Zhejiang Key Laboratory of Medical Imaging Artificial Intelligence, Hangzhou 310058, China.}
\thanks{$^\dag$Equal contribution.}
\thanks{$^*$Corresponding Authors: Haochao Ying and Mingchuan Zhou.}%
}%

\begin{document}
	
\maketitle
\thispagestyle{empty}
\pagestyle{empty}
\begin{abstract}
    
Intraocular foreign body removal demands millimeter-level precision in confined intraocular spaces, yet existing robotic systems predominantly rely on manual teleoperation with steep learning curves.    To address the challenges of autonomous manipulation—particularly kinematic uncertainties from variable motion scaling and Remote Center of Motion (RCM) point variation—we propose \model, an imitation learning framework for autonomous intraocular foreign body ring manipulation.    Our approach integrates RCM dynamic calibration to resolve coordinate system inconsistencies caused by intraocular instrument variation and introduces the RCM-ACT architecture, which combines action chunking transformers with episode-level kinematic realignment.    Trained solely on stereo visual data and instrument kinematics from expert demonstrations in an artificial eye model, \model successfully completes ring grasping and positioning tasks without explicit depth sensing. Experimental validation demonstrates the successful implementation of end-to-end autonomy under uncalibrated microscopy conditions\add{, achieving a mean 3-D Euclidean grasp deviation of 0.686~mm and 11/20 full-task successes}. The results provide a viable framework for developing intelligent eye surgical systems capable of complex intraocular procedures.

\end{abstract}

\section{INTRODUCTION}



Intraocular foreign body (IOFB) removal represents a critical clinical procedure in vitreoretinal surgery. IOFBs can originate from diverse sources—metallic particles from grinding or welding accidents, glass fragments from ocular trauma, and ceramic or stone materials from occupational injuries—each presenting distinct mechanical and imaging challenges. The clinical urgency is particularly acute, as retained IOFBs can cause chronic inflammation, progressive retinal damage, and potential vision loss if not removed promptly and safely. Standard manual removal involves complex coordination of microscopic visualization, delicate instrument manipulation around vital retinal structures, and precise grasping under high magnification, making the procedure demanding even for expert surgeons.

Intraocular foreign body removal requires submillimeter precision to safely remove fragments near delicate retinal tissues while minimizing iatrogenic damage\cite{wirostko2019removal,al2024clinical,liu2023clinical,jung2021intraocular}. 
Robotic assistance has shown great promise in enhanced control and stability to reduce physiological tremors\cite{singh2022overcoming,retnaningsih2023robotics,thirunavukarasu2024robot}.
However, current systems still rely on manual teleoperation, which introduces steep learning curves and operational inefficiencies.
While automated removal offers the potential for consistent high-precision execution, key challenges include accurately identifying optimal grasping points (pose determination for action initiation) and continuously evaluating fragment adhesion states in real time (action status assessment and adaptive path planning), which have so far impeded the deployment of automated solutions.

Recent advances in large-scale imitation learning have significantly advanced the automation of robotic manipulation tasks~\cite{ingelhag2025real,wang2025deri,zhang2025multi,imai2025autonomous}.
 For instance, imitation learning was used for two Franka robots employed in the peg transfer task in the fundamentals of laparoscopic surgery~\cite{kawaharazuka2024robotic} and has also been applied to path planning in orthopedic surgeries~\cite{jian2025motion}.
However, the application of imitation learning in micro-scale eye robotic systems has not been widely studied~\cite {marinho2025evaluation,arikan2024towards,kim2021towards}.
Such systems are typically equipped with micromanipulators and advanced intraocular imaging capabilities~\cite{briel2025intraoperative}, generating rich multimodal datasets during procedures.
These datasets include stereomicroscopic video streams and instrument kinematics.
Leveraging this data enables the training of autonomous control policies for eye microsurgical tasks.
In this paper, we explore intraocular foreign body removal operation, the complex yet representative task, to assess the potential of imitation learning in ophthalmic procedures.


However, applying imitation learning to eye surgical robot introduces two distinct technical challenges not commonly encountered in other robotic applications:
1) \textit{Variable Motion Scaling.} Intraocular foreign body removal systems allow surgeons to adjust the control-to-motion ratio to transition between gross movements and fine manipulations. This variability introduces kinematic uncertainty, as the same control input may result in different instrument motions depending on the scaling factor.
Consequently, imitation learning policies trained on raw control inputs without accounting for these scaling variations may fail to generalize effectively during actual surgical procedures. 
2) \textit{Remote Center of Motion Variation.}
In ophthalmic robotic surgery, instruments are introduced into the eye through a fixed entry point in the sclera, known as the Remote Center of Motion (RCM). Maintaining precise instrument motion around this pivot is crucial to prevent tissue damage. However, slight positional deviations or variations in the RCM point can disrupt the spatial coordinate system, leading to inaccuracies in instrument positioning. Unlike traditional robotic systems, recalibration is complex due to anatomical constraints.

In this work, we propose \model, an imitation learning framework for autonomous intraocular foreign body removal surgical manipulation. Our method trains control policies through expert demonstrations of grasping and placing foreign body in an artificial eye model, utilizing solely instrument kinematics and stereo visual data. To address the unavoidable variability in motion scaling inherent to microsurgical operations, we bypass kinematic ambiguity by directly training policies on instrument actuator-level displacements rather than raw control signals, ensuring generalization across magnification levels while remaining extensible to diverse microsurgical platforms. For the critical RCM variation challenge, we devise an episode-level coordinate alignment strategy: realignment of instrument positions against three predefined phantom landmarks converts variation-corrupted kinematic data into a unified global coordinate system via a per-episode rigid transform. This software-defined spatial anchoring mitigates cumulative errors from RCM point deviations without requiring hardware recalibration, preserving submillimeter positioning accuracy. By achieving submillimeter-scale grasp deviation in instrument-ring interactions without explicit depth sensing, our study \add{demonstrates the preliminary technical feasibility of autonomous ring-based IOFB proxy manipulation under controlled phantom-eye conditions}.

Our work makes three primary contributions: (i) An end-to-end imitation-learning framework for autonomous intraocular foreign body removal manipulation using an eye surgical robot---to our knowledge the first such system for this task, demonstrating submillimeter-scale grasp deviation; (ii) A novel RCM-based state representation resolving motion scaling uncertainties in microsurgical interfaces;  (iii) Successful validation of end-to-end visuomotor policies for intraocular manipulation under uncalibrated microscopy conditions.

\section{RELATED WORK}\label{sec:2}
We analyze two related advanced areas: eye surgical robotic systems and imitation learning for surgical automation, concluding each with identified research gaps.
\subsection{Eye Surgical Robotic Systems}

Eye surgical automation has progressed through three distinct technological phases, each addressing specific limitations of prior systems. Initial first-generation platforms like the IRISS robot achieved 0.2 mm precision through mechanical Remote Center of Motion (RCM) constraints~\cite{rahimy2013robot}, yet necessitated complete human trajectory control due to lacking intraoperative sensing. Second-generation systems overcame this limitation by integrating micron-scale optical coherence tomography (OCT), exemplified by Ebrahimi~et~al.~\cite{ebrahimi2022simultaneous} whose vision-force fusion algorithm enhanced instrument localization accuracy by 37\% without preoperative calibration. Current third-generation solutions demonstrate partial autonomy across three technical dimensions: (1) tissue interaction control via spectral OCT-guided robot reducing retinal tears by 41\% through real-time elastography (10 mN tension detection)~\cite{keller2020optical}, (2) workflow optimization with systems like Preceyes Surgical System (PSS) cutting operative time by 28\%~\cite{turgut2023robot}, and (3) instrument guidance through Zhou~et~al.~\cite{zhou2019towards} hybrid robot achieving 12 µm needle placement accuracy via software RCM control, complemented by their 98.7\% accurate deep learning-based needle detection~\cite{zhou2023needle}. However, these advancements remain fragmented—no system integrates all three dimensions simultaneously, existing autonomy operates in isolated surgical phases, and clinical validation is confined to porcine models. This fragmentation motivates our investigation into \del{cross-modality integration}\add{a unified visuomotor framework that fuses stereo-microscopic imaging with kinematic feedback within a single end-to-end policy} toward establishing an end-to-end automation framework for human \del{clinical applications}\add{intraocular foreign body removal under uncalibrated microscopy---to our knowledge the first such system for this task. The present prototype uses stereo microscopy and robot proprioception as inputs.}

\addblk{Complementary to fully autonomous systems, shared-control and partially autonomous architectures---including co-manipulated micromanipulators with virtual-fixture safety layers---combine surgeon judgement with robotic precision and represent another major line of intraocular robotic research. Recent reports further caution that fully robotic execution can prolong overall procedure time relative to manual surgery~\cite{eberle2024surgical}. In parallel, recent updates from the GEYEDANCE programme have reported human-eye anatomical-model evaluation of vitreoretinal robotic manipulation~\cite{piccinelli2025geyedance}, thereby narrowing the gap between porcine-only validation and clinical translation.}

\subsection{Imitation Learning in Robotics}

Imitation learning has been widely explored in robotic manipulation tasks, enabling robot to learn from expert demonstrations by replicating human actions. A common method in early research was behavioral cloning, which predicts actions directly from observations. As deep learning and generative modeling advanced, new architectures like ConvNets and Vision Transformers (ViTs)\cite{dosovitskiy2020image} were employed for image processing, while recurrent networks and transformers\cite{kim2021transformer,wang2022transformer} helped incorporate temporal dependencies in the observations. In addition, generative approaches, including energy-based models, diffusion models, and VAEs, were introduced to better capture complex motion patterns from demonstrations~\cite{katara2024gen2sim,pang2021trajectory,wang2024sparse,chen2024semantically}. \textcolor{red}{Among these advancements, Action Chunking with Transformers (ACT)~\cite{zhao2023learning} has emerged as a landmark framework for learning complex, long-horizon manipulation tasks from demonstrations. Diverging from traditional behavioral cloning, which predicts a single action at each timestep, the core idea of ACT is to predict a sequence of future actions, or an ``action chunk.'' This approach leverages the powerful self-attention mechanism of Transformers to capture long-term temporal dependencies in the observation data, resulting in smoother and more coherent robot trajectories. While ACT has demonstrated remarkable success in tabletop manipulation tasks within structured environments, its direct application to microsurgical domains like intraocular procedures reveals a key limitation: its reliance on a consistent coordinate frame. The unique challenge of a shifting Remote Center of Motion (RCM) in ophthalmic surgery, absent in typical robotic setups, compromises the precision of the standard ACT framework and motivates the adaptations presented in this work.}

Although end-to-end imitation learning theoretically holds the potential to address the complexities of surgical tasks, its effective application in the surgical domain remains an open question.  Recent efforts have begun to apply imitation learning in surgical contexts~\cite {marinho2025evaluation,arikan2024towards,kim2021towards}, such as Kim et al.\cite{kim2024surgical}, which introduced the Surgical Robot Transformer (SRT) for surgical tasks using approximate kinematics data from the da Vinci system, overcoming challenges related to inaccurate proprioception and hysteresis. However, this approach is not applicable in intraocular foreign body removal surgery, as such procedures require precise contact with real-world objects and accurate localization. Moreover, challenges persist in applying imitation learning to fine-scale surgical tasks, particularly those requiring variable-scale manipulation.

\begin{figure*}[t]
\vspace{0.13cm}
\centering
    \includegraphics[width=0.8\textwidth]{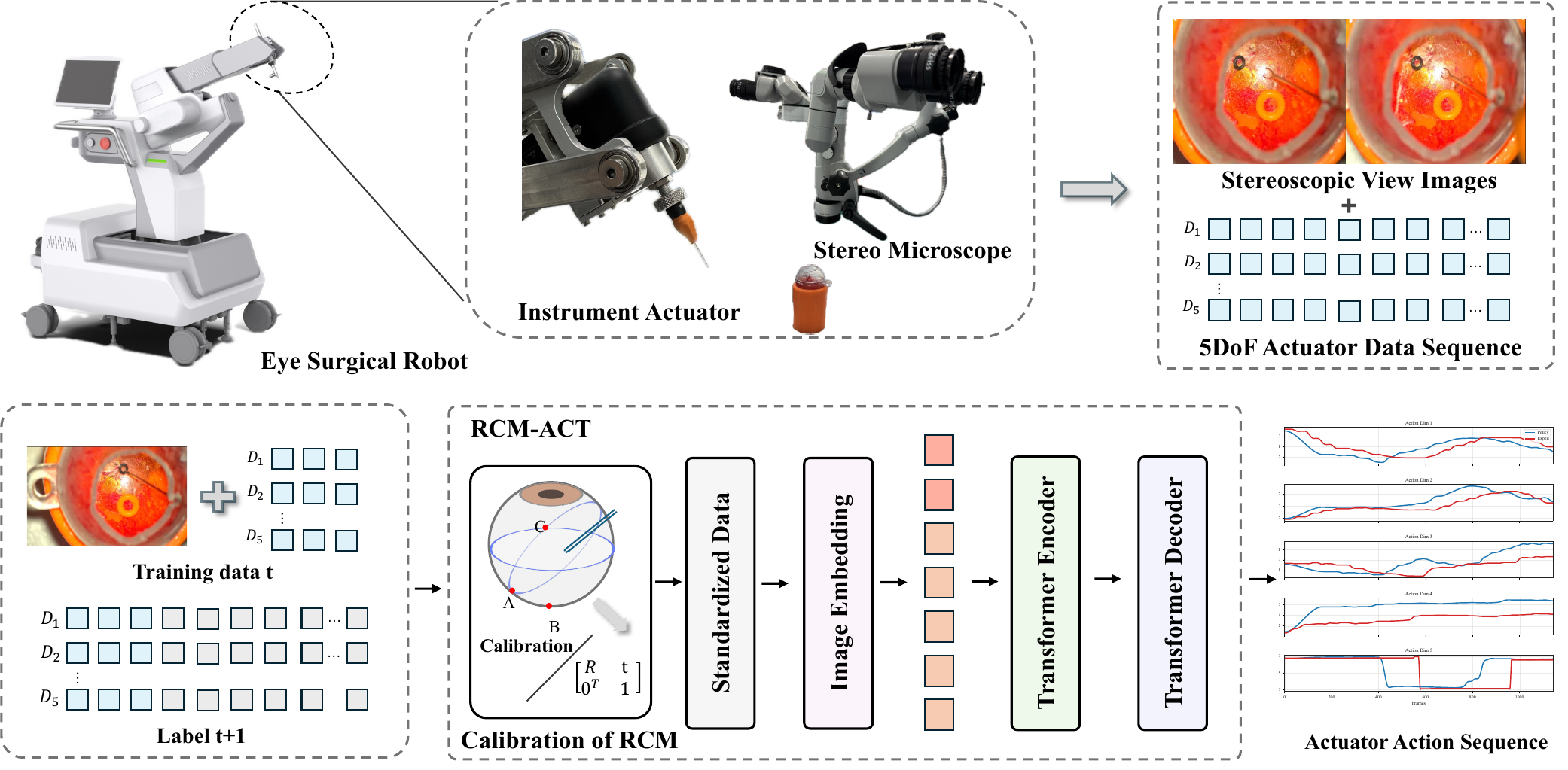}
    \caption{Overview of RCM-ACT Framework: Instrument Actuation with Data Acquisition and \del{Multimodal}\add{Stereo-Microscopic} Training. \add{Training inputs are stereo microscopy and robot proprioception. Blue and red curves in the ``Actuator Action Sequence'' panel correspond to policy-predicted and expert trajectories.}}
    \label{fig:ramework:}
\end{figure*}

\section{METHOD}
\label{sec:3}
The RCM-ACT framework, illustrated in Figure~\ref{fig:ramework:}, combines per-episode RCM calibration with an action-chunking transformer trained on stereo microscopy and robot proprioception. The following subsections detail the task formulation, the RCM dynamic calibration, the RCM-ACT architecture, and the inference procedure.

\subsection{Task Definition}
We use the classic ring grasp-and-place operation to simulate intraocular foreign body removal\cite{lin2020low}. \add{The ring pair is a deliberate IOFB proxy: the annular geometry provides sub-millimetre alignment ground-truth absent from irregular fragments, the black-on-orange contrast replicates surgical visual saliency, and the threading sub-task enforces the same 6-DoF pose commitment that dominates IOFB retrieval. Grasping accuracy is the primary metric, and placement is evaluated as a coupled secondary metric.} A robotic manipulator is used to manipulate a 3D eye model $\mathcal{M}$, consisting of two rings: a black ring $R_{\text{black}}$ and a larger orange ring $R_{\text{orange}}$. The task involves automatically grasping the black ring $R_{\text{black}}$ and placing it onto the orange ring $R_{\text{orange}}$ with high precision. At each timestep $t$, the robot receives an observation $o_t$, which includes camera images of the model  $\mathcal{M}$, showing the positions of the rings
$\mathbf{p}_{R_{\text{black}}}[t]$ and $\mathbf{p}_{R_{\text{orange}}}[t]$, as well as the robot’s end-effector pose $\mathbf{p}_t = (x_t, y_t, z_t)$ and orientation $\mathbf{r}_t$. The robot also receives proprioception \add{$\mathbf{q}_t$, where we reserve the scalar $x_t$ for the Cartesian $x$-coordinate of the end-effector while $\mathbf{q}_t = [x_t,y_t,z_t,\theta_t,\dot{g}_t]^\top$ denotes the full proprioceptive arm-pose vector}. Based on these inputs, the robot computes an action $a_t$ to move the end-effector to grasp the black ring $R_{\text{black}}$ and place it on the orange ring $R_{\text{orange}}$. The task is learned from demonstrations, where each trajectory $\tau = \{(o_1, \mathbf{q}_1, a_1), \dots, (o_T, \mathbf{q}_T, a_T)\}$ consists of a sequence of observations, proprioception, and actions. The robot aims to minimize misalignment and collision, ensuring precise placement of the black ring on the orange ring.
	
	\subsection{RCM Dynamic Calibration}

In each task demonstration, during the data acquisition process, there exists a shift in the position of the point of insertion into the eye, causing a deviation in the position of the Remote Center of Motion (RCM). This results in a variation of the RCM point, meaning that the coordinate system for each data collection is inconsistent. To address this issue, we propose a dynamic calibration method for the RCM. \textcolor{red}{RCM dynamic calibration is only performed when the RCM point varies or shifts. This variation typically occurs when the instrument is restarted or upon its entry into the eye, which can cause a shift in the point of insertion. The calibration is completed once at the beginning of each complete task episode, such as a full demonstration or a single autonomous execution, rather than continuously at every timestep.}

The dynamic calibration is achieved by using the robotic instrument to reach three fixed points $\mathbf{p}_1, \mathbf{p}_2, \mathbf{p}_3$ within the workspace. \add{In our artificial-eye setup, these three points are pre-machined dimples on the inner wall of the eye phantom---they are not placed on the retinal surface of the phantom and are revisited only by the gripper tip.} By obtaining the positions of these three fixed points in the episode-start coordinate system, we compute the corresponding rigid transform $\mathbf{R}_\tau$ that aligns the local coordinate system to a global one. This transform allows the collected data from varying coordinate frames to be remapped into a common reference frame. \textcolor{red}{The process of reaching the three fiducial markers for calibration was performed under the same RCM constraint as the surgical manipulation itself. The robot's motion was pivoted around the established RCM point at the eye's entry, ensuring that the calibration procedure accurately reflects the kinematics of the actual task.}

Formally, once per episode the instrument visits three non-coplanar fiducial points $\mathbf{p}_1, \mathbf{p}_2, \mathbf{p}_3$ in the current frame $\mathcal{C}_t$:
\begin{equation}
    p_i(\tau) = \left[ x_i(\tau),\, y_i(\tau),\, z_i(\tau) \right]^T, \quad i = 1, 2, 3.
\end{equation}
A rigid transform $(R_\tau, d_\tau)$ is then recovered by solving
\begin{equation}
    p_i(\tau) = R_\tau \cdot p_i(0) + d_\tau, \quad i = 1, 2, 3,
\end{equation}
where $p_i(0)$ is the reference position in $\mathcal{C}_0$. All subsequent points are remapped via
\begin{equation}
    p' = R_\tau^T (p - d_\tau).
\end{equation}

By applying the episode-level transform $(R_\tau, d_\tau)$ to each collected point, all demonstrations are mapped into a consistent coordinate system for subsequent policy learning.

\begin{figure*}[t]
    \centering
    \begin{subfigure}[b]{0.43\linewidth}
        \centering
        \includegraphics[width=\linewidth]{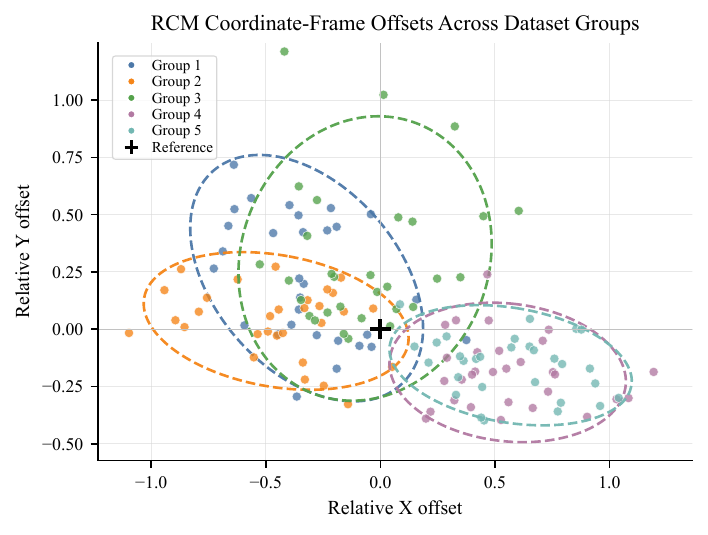}
        \caption{\addblk{Episode-start RCM frame offsets by dataset group.}}
        \label{fig:rcm_dataset_offsets}
    \end{subfigure}
    \begin{subfigure}[b]{0.43\linewidth}
        \centering
        \includegraphics[width=\linewidth]{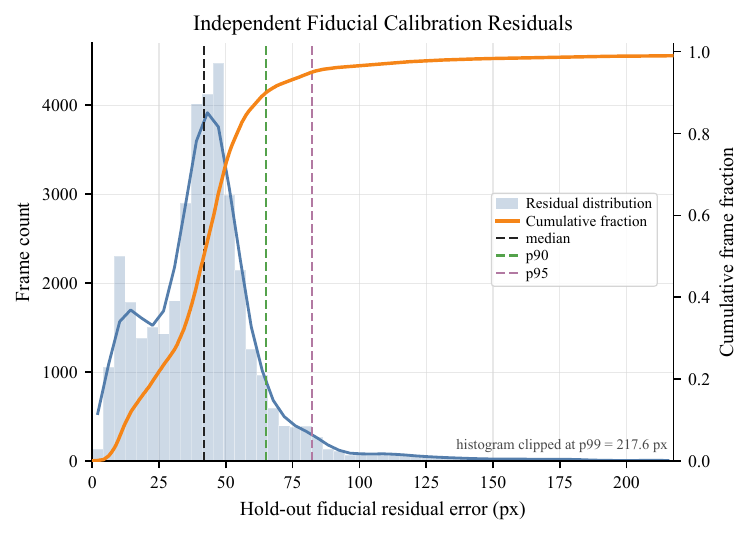}
        \caption{\addblk{Hold-out fiducial residual.}}
        \label{fig:fiducial_residuals}
    \end{subfigure}
    \caption{\addblk{RCM dynamic calibration analysis. \textbf{(a)}~Episode-start coordinate-frame offsets after global-reference centering, separated by dataset group. \textbf{(b)}~Calibration residual evaluated with a held-out third fiducial: median 41.7~px and 95th percentile 82.4~px ($\approx$20.7\% of marker baseline).}}
    \label{fig:rcm_calib}
\end{figure*}

	\subsection{RCM-ACT Architecture}

\begin{algorithm}[t]
\SetKwInOut{Input}{Input}
\SetKwInOut{Output}{Output}
\Input{Raw demonstration dataset $\mathcal{D} = \{\tau_1, \tau_2, \dots\}$, chunk size $k$, weight $\beta$}
\Output{Updated model parameters $\theta$, $\phi$}
\textbf{Phase 1: Dataset Calibration} \\
\For{each episode $\tau$ in $\mathcal{D}$}{
    Compute rotation matrix $R_{\tau}$ from three fiducial points\;
    \For{each timestep $t$ in $\tau$}{
        Transform $\mathbf{q}_t$ to $\mathcal{C}_0$ using $(R_{\tau}, d_{\tau})$ and pair it with the stereo observation $o_t$\;
    }
}
Let $\mathcal{D}_{\text{calibrated}}$ be the realigned dataset\;
\textbf{Phase 2: Model Training} \\
Initialize encoder $q_\phi(z|a_{t:t+k}, \overline{o_t})$ and decoder $\pi_\theta(\hat{a}_{t:t+k}|o_t, z, \mathcal{C}_0)$\;
\For{each training step}{
    Sample $(o_{t}, a_{t:t+k})$ from $\mathcal{D}_{\text{calibrated}}$; sample $z \sim q_\phi$\;
    Compute $L = \text{MSE}(\hat{a}_{t:t+k}, a_{t:t+k}) + \beta\, D_{\text{KL}}(q_\phi \parallel \mathcal{N}(0,I))$\;
    Update $\theta, \phi$ with AdamW\;
}

\caption{RCM-ACT Training with Calibration}
\label{alg:rcm_act_training_revised}
\end{algorithm}

\textcolor{red}{The standard ACT framework predicts future action chunks under the assumption of a stable robot motion base, but the entry-point-induced RCM variation in ophthalmic surgery violates this assumption: an unmodified ACT policy trained on raw kinematics conflates intra-episode end-effector motion with episode-level coordinate-frame shifts. RCM-ACT addresses this by tightly integrating the dynamic calibration of Section~III-B with the ACT pipeline. As detailed in Algorithm~\ref{alg:rcm_act_training_revised}, kinematic data from every demonstration episode is first transformed into the unified coordinate system $\mathcal{C}_0$ via the per-episode rigid transform $(R_\tau, d_\tau)$, and only then is the action-chunking transformer trained on the realigned trajectories. The policy therefore learns a single motion distribution under one consistent reference frame, rather than a mixture over RCM-shifted frames.}

\textcolor{red}{Two design choices distinguish RCM-ACT from a vanilla ACT trained on raw kinematics. First, the encoder $q_\phi(z|a_{t:t+k}, \overline{o_t})$ and decoder $\pi_\theta(\hat{a}_{t:t+k}|o_t, z, \mathcal{C}_0)$ both operate in the realigned frame $\mathcal{C}_0$, so the latent variable $z$ captures only intra-episode motion variability rather than entangling it with inter-episode coordinate-frame shifts. Second, the per-frame proprioception $\mathbf{q}_t$ supplied to the policy is read from forward kinematics after the realignment transform has been applied, which decouples the policy input from any controller-side teleoperation scaling that varied across demonstrations. These two choices together explain why the gain over baseline ACT (Table~\ref{tab:act_rcm_train}) cannot be attributed to architectural changes alone---the realignment of the training distribution is the load-bearing component.}

The resulting policy learns action chunks in the realigned frame $\mathcal{C}_0$, and the predicted actions are executed by the robot controller under the software-enforced RCM constraint, which is maintained at the controller level. \add{The proposed calibration removes cross-episode coordinate-frame inconsistency before policy learning, so the transformer models local manipulation dynamics in a common reference frame rather than absorbing episode-dependent coordinate shifts into the policy.}

To train the RCM-ACT model for a new task, we begin by collecting human demonstrations via rostopic for data transmission and dataset recording. \textcolor{red}{The action $a_t$ at each timestep is defined as the target state vector for the robot's end-effector, represented by $a_t = [x, y, z, \theta, \dot{g}]^\top$, where $[x, y, z]$ represents the Cartesian coordinates, $\theta$ is the gripper's rotation angle, and $\dot{g}$ denotes the gripper's opening/closing state. This target state is then translated into low-level actuator commands by the robot's internal controller. Instead of mapping joystick or controller inputs to robot actions---a mapping that is directly affected by the scaling factor---our policy learns a direct mapping from the current state (composed of stereo visual data and instrument kinematics) to the desired next state (represented as instrument actuator-level displacements).} The input to the system includes stereo imagery from a stereo microscope, which provides real-time visual feedback from two camera views. In addition to the visual data, the input also contains values representing the Cartesian coordinates of the current position and the rotation encoder value of the gripper. These proprioceptive and visual observations are essential for the task as they offer real-time sensory data about the robot's state and environment.

Once the data is collected, the RCM-ACT model is trained to predict future actions based on the current observations. The predicted actions correspond to the target joint positions for the robotic arm at the next timestep, mimicking how a human operator would proceed given the current observations. These predicted target joint positions are then used to guide the robotic arm in achieving the desired configurations. Concretely, the ACT module consists of an encoder $q_\phi(z|a_{t:t+k}, \overline{o_t})$ that learns a latent variable $z$ conditioned on the action sequence $a_{t:t+k}$ and the observation sequence $\overline{o_t}$ (the observation without image data), and a decoder $\pi_\theta(\hat{a}_{t:t+k}|o_t, z, \mathcal{C}_0)$ that generates the predicted future actions $\hat{a}_{t:t+k}$ based on the current observations $o_t$, the latent variable $z$, and the global coordinate system $\mathcal{C}_0$. The model is trained to minimize the reconstruction loss $L_{\text{reconst}} = \text{MSE}(\hat{a}_{t:t+k}, a_{t:t+k})$ together with the Kullback--Leibler regularization loss $L_{\text{reg}} = D_{\text{KL}}(q_\phi(z|a_{t:t+k}, \overline{o_t}) \parallel \mathcal{N}(0, I))$ that constrains the latent space to follow a standard Gaussian distribution. The integration of RCM dynamic calibration with the ACT pipeline is therefore essential for achieving smooth, precise, and reliable robotic motion in the presence of inter-episode coordinate-frame variation.

\subsection{Inferring phase}


In this section, we describe the inference process utilized in the robot-assisted task for foreign body ring placement, where the robotic manipulator is responsible for automatically picking up a black ring and placing it inside a larger orange ring within a scaled eye model. This process involves acquiring data from the robot's sensors, performing the episode-start RCM calibration, and predicting actions that guide the robot in completing the task.

At the start of each episode, the episode-start RCM calibration (Section~III-B) computes the per-episode rigid transform $(R, d)$ once. The robot's position $\mathbf{p}_t = (x_t, y_t, z_t)$ and orientation are then expressed in the unified frame $\mathcal{C}_0$ via $(R, d)$, ensuring that the robot operates within a consistently calibrated coordinate system throughout the episode. The transform is applied to all subsequent proprioceptive states so that the policy input remains aligned with the global frame of reference used during training.

Once calibration is completed, the policy inference loop begins. At each timestep $t$, the robot's observation $o_t$ is obtained from the stereo microscope together with the realigned proprioceptive state $\mathbf{q}'_t$. The trained policy $\pi_\theta$ predicts a $k$-step action chunk $\hat{a}_{t:t+k}$, which is appended to a First-In, First-Out (FIFO) buffer $\mathcal{B}$ for future timesteps. The executed action at each timestep is a weighted average over the stored chunks $a_t = \sum_i w_i\,\mathcal{B}[t][i]/\sum_i w_i$ with $w_i = m^i$ and $m=0.8$, so that more recent predictions exert a stronger influence on the current command.

During inference, the robot's data (proprioception and predicted actions) is normalized using pre-defined training statistics, ensuring that the data fed into the model aligns with the expected range for accurate predictions. After the model predicts the actions, the data is denormalized back to its original scale to ensure that the robot's actions correspond accurately to the real-world environment in which it operates. Finally, the computed action is communicated back to the robot through a rostopic labeled \texttt{/model\_output}, allowing the robot to execute the task of picking up the black ring and placing it inside the orange ring. This process integrates per-episode calibration with action prediction to ensure that the robot performs the task with precision and efficiency. \add{Algorithm~\ref{alg:rcm_act_inference} summarizes this deployment procedure.}

\begin{algorithm}[t]
\caption{\add{RCM-ACT Inference with Episode-Level Realignment}}
\label{alg:rcm_act_inference}
\KwIn{\add{Trained policy $\pi_\theta$, episode transform $(R,d)$, stereo observation stream $o_t$, proprioceptive stream $\mathbf{q}_t$, buffer decay $m$, chunk length $k$}}
\KwOut{\add{Executed robot action $a_t$}}
\add{Initialize FIFO action buffer $\mathcal{B}$\;}
\For{\add{each timestep $t$}}{
    \add{Acquire stereo observation $o_t$ and proprioceptive state $\mathbf{q}_t$\;}
    \add{Transform $\mathbf{q}_t$ to $\mathcal{C}_0$ as $\mathbf{q}'_t$ using the fixed episode transform $(R,d)$\;}
    \add{Predict action chunk $\hat{a}_{t:t+k} \leftarrow \pi_\theta(o_t, z=0, \mathcal{C}_0)$ using the realigned proprioceptive state $\mathbf{q}'_t$\;}
    \add{Insert $\hat{a}_{t:t+k}$ into the FIFO buffer $\mathcal{B}$\;}
    \add{Compute $a_t=\sum_i w_i\,\mathcal{B}[t][i]/\sum_i w_i$, where $w_i = m^i$ and $m=0.8$\;}
    \add{Denormalize and send $a_t$ to the robot controller\;}
}
\end{algorithm}

\section{EXPERIMENTS}
\label{sec:4}

\subsection{Experiment Setup}
\subsubsection{Task}

The robotic system was tasked with autonomously manipulating foreign body rings in a simulated intraocular environment, requiring precise grasping of a 1.22 mm black ring (\(R_{\text{black}}\)) and threading it onto a 4.02 mm orange ring (\(R_{\text{orange}}\)). The task specifications included positioning accuracy \(< 0.15\) mm and an operation time limit of \(20 \pm 5\) s. \add{The 0.15~mm tolerance follows from the one-sided clearance between the 1.22~mm ring inner diameter and the 0.92~mm gripper-jaw opening at the grasp pose, beyond which the jaw tips would clip the ring rim. The $20\pm 5$~s limit was set from the completion-time distribution of the 30 expert demonstrations.} The eye phantom was mounted on a 7-DoF motion platform to simulate anatomical variations during ophthalmic surgery. \add{The underlying robot is a custom 6-DoF intraocular micromanipulator with a software-enforced RCM (rated repeatability $\pm 25~\mu$m at the tool tip), driven through a ROS Noetic stack at 60~Hz. The control loop uses a 5~Hz policy thread and a 15~Hz thread applying the precomputed RCM realignment transform, with an inference-and-communication delay of $\le 80$~ms excluding the 5~Hz policy update period.}

\subsubsection{Data Collection}
We collected 30 demonstration episodes ($112 \pm 18$\,s each) with randomized ring positions \add{from a board-certified vitreoretinal surgeon with more than 10 years of intraocular-surgery experience} (referred to as the ``expert'' throughout the paper), generating 112 GB of training data. A motion scaling factor of 1/30 was applied during the expert demonstrations, mapping the operator's hand movements from the controller to the instrument's motion. Data acquisition utilized ROS-based pipelines with asynchronous recording:
\begin{itemize}
    \item Proprioceptive data (5D end-effector state $\mathbf{q}_t = [x,y,z,\theta,\dot{g}]^\top \in {R}^5$) sampled at 60 Hz via rostopic
    \item Stereo vision streams (1920$\times$2160@60 Hz) captured using stereo microscope cameras
    \item Raw sensor data stored in HDF5 format without temporal alignment
\end{itemize}
\add{Although teleoperation scaling factors may vary between demonstrations, RCM-ACT operates on robot-frame proprioceptive states $\mathbf{q}_t$ obtained from forward kinematics rather than raw controller commands. Thus, the scaling factor does not enter the policy input or output, avoiding direct dependence on controller-side scaling variability.}

\subsubsection{Training of Models}

The RCM-ACT model was trained for 2000 epochs on an NVIDIA RTX 4090 GPU using AdamW optimizer ($\alpha=3\times10^{-4}$, $\beta_1=0.9$, $\beta_2=0.95$) with cosine learning rate decay. Training configurations included: 32-sequence batches (8 timesteps each), dropout rate 0.1, and KL-divergence regularization ($\beta=0.5$). Model selection used five independent train/test splits derived from the 30 demonstration episodes, with chunk size 90 chosen via the sensitivity analysis in Section~IV-C. The KL-divergence weight $\beta=0.5$ was chosen to balance reconstruction fidelity and latent-space regularisation.

\subsubsection{Deployment and Inference of Model}
\begin{figure*}[t]
    \vspace{0.13cm}
    \begin{center}
    \addtolength{\tabcolsep}{-5pt}    
    \renewcommand{\arraystretch}{0.1}
    
       
        \includegraphics[width=0.8\linewidth]{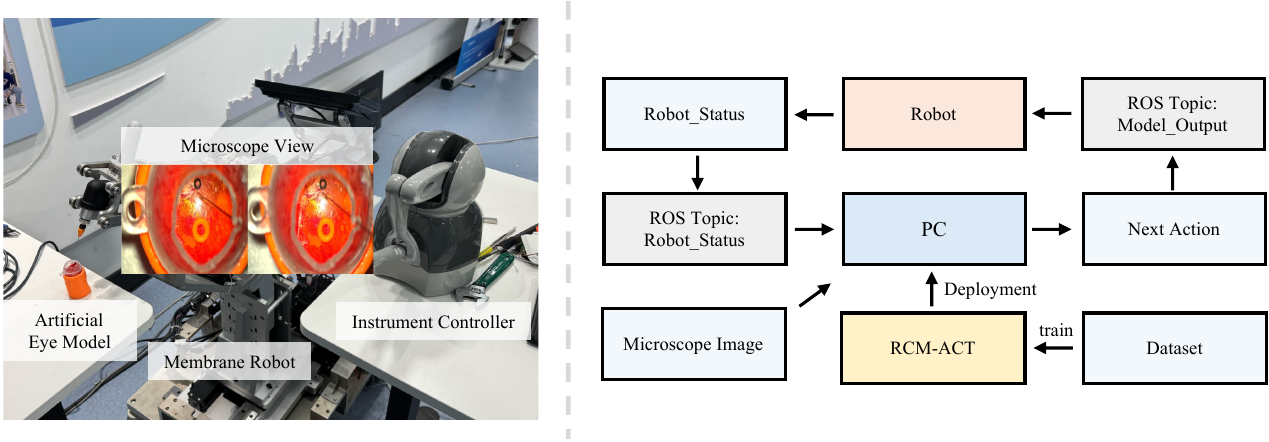} \\

    \caption{Deployment process of the intraocular robot in an artificial eye model environment (3-D printed phantom on a 7-DoF motion stage).}
    \label{fig:Deployment Process}
    \end{center}
\end{figure*}

Figure \ref{fig:Deployment Process} illustrates the actual deployment environment and workflow diagram. Real-time inference was deployed on a mobile workstation (Intel Core\textsuperscript{\textregistered} i9-13900HX, NVIDIA\textsuperscript{\textregistered} RTX 4090 Laptop GPU) running ROS Noetic, where the trained policy generated action sequences ($x$, $ y$, $z$, $\theta$, $\dot{g}^\top $) at 5 Hz through the \texttt{/model\_output} rostopic. A dual-thread architecture achieved parallel execution: 1) \textit{Perception Thread} asynchronously processed stereo images (OpenCV 4.5.5) and proprioceptive states at 30 Hz, 2) \textit{Control Thread} applied the precomputed RCM realignment transform at 15 Hz for action prediction. Temporal misalignment from unsynchronized training data was mitigated by a weighted action buffer employing exponential decay:
\begin{equation}
    \mathbf{a}_t = \sum_{k=0}^{N} m^k \mathbf{a}_{t-k} \quad (m=0.8,\ N=3),
\end{equation}

\begin{table*}[t]
\caption{Performance comparison between deployment results and evaluation tests. \add{Deployment metrics are counted over 20 independent trials per method. Orange-ring reach is evaluated independently of grasp success, whereas full-task success requires both grasping the black ring and releasing it inside the orange ring. RCM-ACT achieved a median completion time of 18.6~s (interquartile range 17.4--20.1~s) over its 11 successful trials, within the $20\pm5$~s operating budget.}}
\centering
\begin{tabular}{c c c c c c c c}
\hline
\multirow{2}{*}{Method}  & \multicolumn{3}{c}{\textbf{Deployment}} & &\multicolumn{3}{c}{\textbf{ Model Evaluation}} \\ 
 & $R_{\text{black}}$ Grasp & $R_{\text{orange}}$ Reach & Full-task Success & & MSE & Grasp Deviation (mm) & Grasping Latency (frame) \\ \hline
 
ACT & 2&5 & 0/20& &0.019876 & 1.182&48.50 \\ 
RCM-ACT w.Encoder &- &- &- & &0.017307 &- &102.00 \\ 
RCM-ACT w/o Resample &3 & 8&0/20  &&0.022512 & 1.386& 94.25\\ 
RCM-ACT &11 & 15& 11/20 & &0.014988 &0.686 & 17.50\\ \hline

\end{tabular}

\label{tab:act_rcm_train}
\end{table*}

		
	
\subsection{Experiment Results}
\textcolor{red}{To quantitatively evaluate the policy's performance, we use the following metrics, reported in Table~\ref{tab:act_rcm_train}.} \add{Robot trajectories are compared against expert demonstrations, which provide a single repeatable reference standard for both grasp-frame alignment and timing metrics.}

Three metrics are reported in Table~\ref{tab:act_rcm_train}:

\begin{itemize}
    \item \textbf{Mean Squared Error (MSE):} This metric measures the fidelity of the policy's predicted action trajectory to the expert's ground-truth trajectory on a held-out test set. It is calculated as
    \begin{equation}
        \text{MSE} = \frac{1}{T} \sum_{t=1}^{T} \lVert a_{\text{predicted}, t} - a_{\text{expert}, t} \rVert_2^2,
    \end{equation}
    where $a_t$ is the 5D action vector $[x, y, z, \theta, \dot{g}]^\top$ at timestep $t$ and $T$ is the total number of timesteps. The expert trajectory is used strictly for evaluation and is not available to the policy during inference.

    \item \textbf{Grasp Deviation:} This measures the positional accuracy at the critical moment of grasping the ring. It is defined as the 3-D Euclidean distance between the autonomous agent's gripper and the expert's gripper at the frame of grasping:
    \begin{equation}
        \text{Grasp Deviation} = \lVert p_{\text{robot, grasp}} - p_{\text{expert, grasp}} \rVert_2,
    \end{equation}
    where $p_{\text{grasp}}$ is the 3-D coordinate vector $[x, y, z]$ of the end-effector when the grasp is initiated.

    \item \textbf{Grasping Latency:} This metric evaluates how well the policy replicates the timing of the expert's actions. It is the absolute difference in frame number (timestep) between when the autonomous agent initiates the grasp versus when the expert did in the demonstration:
    \begin{equation}
        \text{Grasping Latency} = |t_{\text{robot, grasp}} - t_{\text{expert, grasp}}|,
    \end{equation}
    where $t_{\text{grasp}}$ is the timestep at which the grasping action occurs.
\end{itemize}

\begin{figure}[t]
    \centering
    \includegraphics[width=\linewidth]{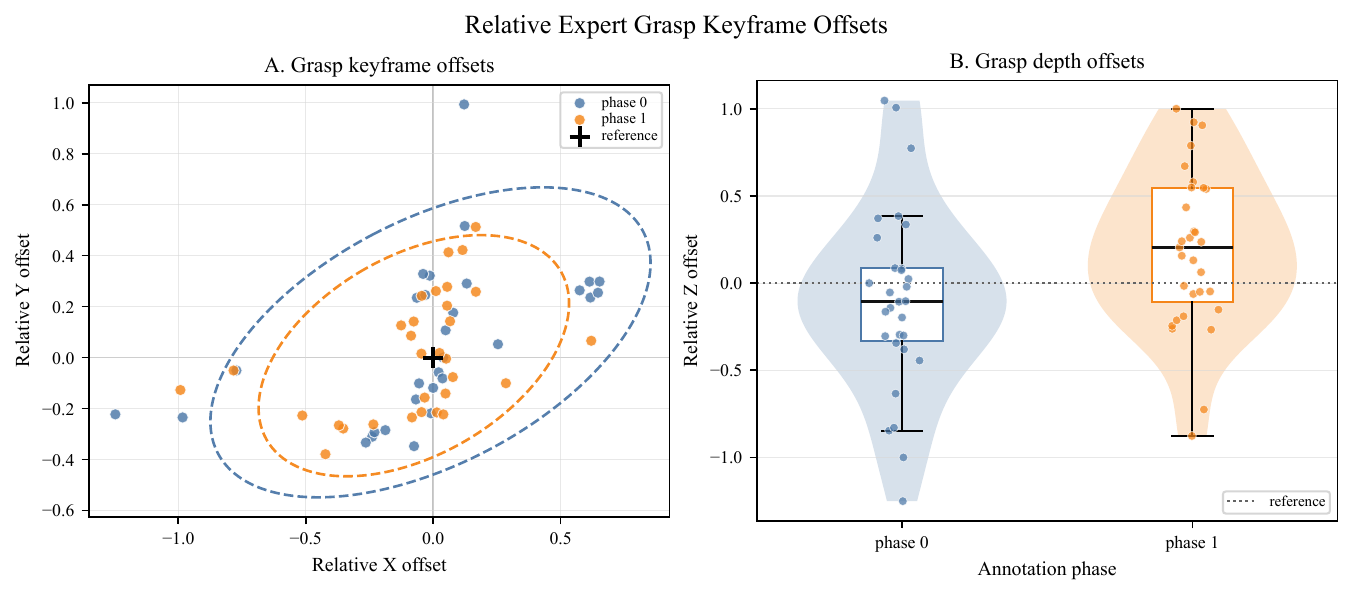}
    \caption{\addblk{Spatial variability of expert grasp keyframes. Keyframes are centered and normalized; axes indicate relative offsets. Left: planar offsets with phase-wise covariance ellipses. Right: depth-offset distributions per phase.}}
    \label{fig:expert_grasp_keyframes}
\end{figure}

\add{Figure~\ref{fig:expert_grasp_keyframes} illustrates the spatial repeatability of the expert grasp keyframes underlying these grasp-frame metrics.}

The results in Table \ref{tab:act_rcm_train} reveal significant performance variations across methods in both deployment and evaluation metrics, emphasizing the critical role of dynamic RCM calibration and resampling mechanisms. \add{The encoder variant is shown only for offline evaluation. Deployment column definitions follow the Table~\ref{tab:act_rcm_train} caption.} In deployment tests, the baseline ACT method~\cite{zhao2023learning} achieved partial success, grasping the black ring twice and \del{placing the orange ring five times across five trials}\add{reaching the orange ring 5 times across the 20 trials}, but failed to complete the full task sequence due to insufficient coordination between subtasks.  The RCM-ACT variant without resampling performed marginally better, with \del{three successful grasp and egiht placements}\add{3 successful grasps and 8 reaches}, yet also failed to achieve full-task completion.  In contrast, the full RCM-ACT model achieved the best deployment performance among the evaluated methods, with \add{11 black-ring grasps, 15 orange-ring reaches, and 11 full-task successes out of 20 trials (55\% strict joint success)}.  The RCM-ACT with encoder variant could not be deployed effectively, as encoder-induced noise caused excessive mechanical jitter, rendering it inoperable for practical execution.
For evaluation metrics, the full RCM-ACT model outperformed all variants, achieving the lowest mean squared error of 0.014988, indicating precise alignment with expert trajectories.  This result significantly surpassed the baseline ACT and the non-resampling RCM-ACT variant, which exhibited higher errors of 0.019876 and 0.022512, respectively.  Grasp deviation further validated the advantage of RCM-ACT, decreasing to 0.686~mm compared with 1.182~mm for ACT and 1.386~mm for RCM-ACT without resampling. \add{Decomposed by axis, the in-plane (xy) component is 0.12~mm and the depth (z) component is 0.675~mm, indicating that the depth axis dominates the residual error---consistent with the depth-ambiguity failure mode identified in Section~V.}  Latency analysis revealed that RCM-ACT operated with near-real-time responsiveness at 17.50 frames, while the baseline ACT and the encoder-based variant suffered substantial delays of 48.50 frames and 102.00 frames, respectively.  The extreme latency in the encoder variant underscores the destabilizing effects of noise on real-time performance.

\add{RCM-ACT failed in 9 of 20 autonomous trials. Failure cases were characterized by frame-level inspection of the stereo recordings and grouped into three error categories. Grasp slippage was the dominant failure mode, occurring in five trials, where the gripper closed 0.4--1.1~mm above the ring rim under near-vertical approach angles. Three trials showed lateral mis-reach, with the tip reaching within 0.3--0.8~mm of the ring center but at an approach angle that placed the jaws outside the annulus. These cases coincided with mid-episode RCM drift greater than 0.2~mm, which cannot be corrected by the current once-per-episode calibration procedure. One trial failed due to post-grasp drop caused by a grip-orientation deviation greater than 10$^\circ$.}

\add{Beyond the table, the calibration analysis in Fig.~\ref{fig:rcm_calib} clarifies where deployment errors originate. Panel (a) plots episode-start coordinate-frame offsets across dataset groups, showing the cross-episode RCM variation that the proposed realignment is designed to remove. Panel (b) quantifies the hold-out calibration residual after realignment: median 41.7~px and 95th-percentile 82.4~px ($\approx$20.7\% of the marker baseline) across 37,954 frames over leave-one-out splits of the 30 demonstration episodes. The realignment thus reduces cross-episode coordinate inconsistency to a bounded range, while the image-plane error contributes to the three lateral mis-reach trials, which exhibited residual mid-episode RCM drift that the once-per-episode calibration could not compensate.}

\subsection{Parameter Sensitivity}
    
\begin{figure}[t]
\vspace{0.13cm}
\centering
    \includegraphics[width=0.4\textwidth]{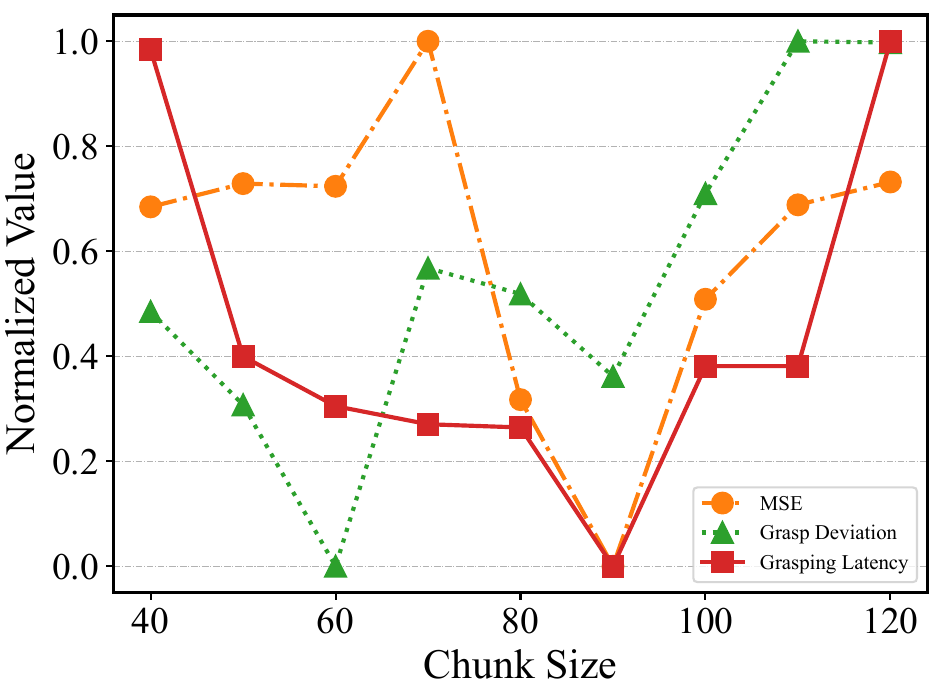}
    \caption{Impact of the Chunk Size}
    \label{fig:chunk size}
\end{figure}

We systematically evaluate the impact of the chunk size parameter on ring task performance in Figure~\ref{fig:chunk size}. To facilitate cross-metric comparisons, all measured values are normalized to the $[0,1]$ range. When increasing the chunk size from smaller to medium values, latency decreases significantly, achieving near-real-time responsiveness at intermediate configurations. Notably, while smaller chunk sizes ($<$90) yield the lowest normalized grasping deviation, their latency remains suboptimal for dynamic robotic operations. Our chosen configuration (chunk size 90) demonstrates an optimal balance, achieving the highest normalized prediction accuracy and near-minimal latency while maintaining grasping deviation within task-specific precision thresholds. Larger chunk sizes ($>$90) exhibit performance degradation across all metrics, with latency increasing by 15--30\% and prediction accuracy decreasing by 12--18\% compared to size 90, indicating overfitting to extended temporal contexts. \add{Beyond chunk size, we further evaluate the fiducial-marker count and the action-buffer parameters.}

\addblk{\textbf{Calibration-point count.} Table~\ref{tab:abl_fiducial} varies $N_{\mathrm{fid}}$ for per-episode realignment. With one marker, the reduced model can only correct translational offset. With two markers, the transform is geometrically underdetermined for full 3-D orientation. With three markers, the system supports an offline leave-one-out residual diagnostic. The numerical trend mirrors this geometric progression: the median residual drops from 138.6 px (one marker) to 76.3 px (two markers) and then to 41.7 px with three, with each step roughly halving the residual. Grasp deviation improves correspondingly from 1.624 mm to 0.987 mm and 0.686 mm, and full-task success rises from 2/20 to 5/20 and 11/20. The largest gain occurs at the transition from two to three markers, where the added rotational redundancy lets the per-episode transform recover the full 3-D pose rather than an axis-ambiguous approximation, which is consistent with the loss of rotational determination and calibration redundancy when fiducials are removed.}

\addblk{\textbf{Action-buffer parameters.} Table~\ref{tab:abl_buffer} varies the exponential-decay weight $m$ and the window length $N$. With a small $m=0.6$, the buffer over-weights stale chunks and latency rises to 31.25 frames while grasp deviation degrades to 0.842 mm and success drops to 7/20. Increasing $m$ to 0.9 reduces latency to 14.75 frames but amplifies chunk-boundary jitter, pushing grasp deviation up to 0.913 mm and success down to 6/20. Widening the window to $N=5$ (with $m=0.8$) keeps grasp deviation low (0.731 mm) but reintroduces latency (25.80 frames) without a success gain. The chosen setting $m\!=\!0.8,~N\!=\!3$ achieves the lowest grasp deviation (0.686 mm), near-minimal latency (17.50 frames), and the highest success rate (11/20), providing the best trade-off among the three metrics.}

\begin{table}[t]
\centering
\caption{\addblk{Ablation on the number of fiducial markers used for per-episode coordinate realignment. Residual statistics are computed from the offline fiducial-residual dataset, whereas grasp deviation and task success are evaluated over 20 deployment episodes per row.}}
\label{tab:abl_fiducial}
\addblk{
\setlength{\tabcolsep}{3pt}
\renewcommand{\arraystretch}{1.15}
\footnotesize
\begin{tabular}{c|cc|cc}
\hline
$N_{\mathrm{fid}}$ & Residual & Residual & Grasp Dev. & Task Success \\
 & median (px) & 95th pct (px) & (mm) & (/20) \\
\hline
1 & 138.6 & 271.4 & 1.624 & 2/20 \\
2 & 76.3  & 154.9 & 0.987 & 5/20 \\
\textbf{3 (ours)} & \textbf{41.7} & \textbf{82.4} & \textbf{0.686} & \textbf{11/20} \\
\hline
\end{tabular}
}
\end{table}

\begin{table}[t]
\centering
\caption{\addblk{Ablation on the action-buffer exponential-decay weight $m$ and window length $N$. All other hyper-parameters match the full RCM-ACT setting in Table~\ref{tab:act_rcm_train}, and metrics are evaluated under the same protocol.}}
\label{tab:abl_buffer}
\addblk{
\setlength{\tabcolsep}{4pt}
\renewcommand{\arraystretch}{1.15}
\footnotesize
\begin{tabular}{cc|ccc}
\hline
$m$ & $N$ & Grasp Dev. (mm) & Latency (frame) & Success (/20) \\
\hline
0.6 & 3 & 0.842 & 31.25 & 7/20 \\
\textbf{0.8} & \textbf{3} & \textbf{0.686} & \textbf{17.50} & \textbf{11/20} \\
0.9 & 3 & 0.913 & 14.75 & 6/20 \\
0.8 & 5 & 0.731 & 25.80 & 9/20 \\
\hline
\end{tabular}
}
\end{table}

\section{DISCUSSION}
\label{sec:5}

\add{The failure distribution suggests clear directions for improvement. Most failures were attributable to depth ambiguity during grasp closure or to residual RCM drift after episode-level calibration, rather than to visual misidentification or coordinate-system collapse. This indicates that the current framework provides a working autonomous prototype for ring-based intraocular manipulation, while further gains are likely to come from explicit depth cues (e.g.~stereo disparity or intraoperative OCT), online RCM updating during the episode, and contact-aware control informed by tool--tissue force feedback. We view the present system as a feasibility study under controlled phantom conditions rather than a clinically deployable solution.}

\addblk{From a safety perspective, post-trial visual inspection after the 20 autonomous trials revealed no damage to the phantom inner surface. The current setup relies on this post-trial visual check and on the software-enforced RCM constraint, without a dedicated contact-event logger, real-time tool--retina clearance monitoring, or a calibrated force-sensing module. Stronger safety guarantees for clinical settings will likely require quantitative tool--phantom-surface contact tracking and force measurements before any in-vivo translation, together with virtual fixtures, action-projection-based no-go zones near the retina, and a real-time tool-to-retina clearance signal from stereo disparity.}

\add{Two clinical challenges remain before routine clinical application. (1)~\emph{Optical distortions}---the real eye's compound optics (cornea, aqueous humor, IOL) are absent in our flat-window phantom, potentially degrading spatial cues during intraocular manipulation. (2)~\emph{Patient-specific RCM}---calibration is per-episode, not per-patient, and the phantom dimples used here would need to be replaced by non-contact surrogates (e.g.\ device-related or vision-based anatomical references) for in-human use. Generalization will further require a brief patient-specific calibration phase or per-patient fine-tuning; identifying and validating clinically repeatable landmarks remains future work.}

\section{CONCLUSIONS}
\label{sec:6}


In this work, we presented RCM-ACT, a framework for autonomous intraocular foreign body removal manipulation using imitation learning with an eye surgical robot. Our approach achieves a 0.686 mm mean 3-D Euclidean grasp deviation and an 11/20 full-task success rate under controlled artificial-eye phantom conditions, by leveraging episode-level RCM dynamic calibration and an action-chunking transformer architecture trained solely on stereo microscopy and robot proprioception, without depth sensing. However, several important limitations remain for clinical translation, including the absence of force sensing, real-time tool--retina clearance monitoring, and per-patient calibration.
Future work will focus on addressing these challenges through explicit depth cues (stereo disparity or intraoperative OCT), force-sensing virtual fixtures, and non-contact patient-specific calibration based on laser-projected spatial references and micro-force feedback at the gripper tip.

	\bigskip

	\bibliographystyle{IEEEtran}
	\bibliography{IEEEexample}

\end{document}